\documentclass[10pt,twocolumn,letterpaper]{article}

\usepackage[pagenumbers]{iccv} 

\usepackage{booktabs}
\usepackage{pifont}
\usepackage{xcolor}
\usepackage{float}
\usepackage{placeins}
\usepackage{booktabs}
\usepackage{multirow}
\usepackage{makecell}
\usepackage{tablefootnote}
\usepackage[accsupp]{axessibility}

\newcommand{\cmark}{\textcolor{green}{\ding{51}}}%
\newcommand{\xmark}{\textcolor{red}{\ding{55}}}%

\definecolor{iccvblue}{rgb}{0.21,0.49,0.74}
\usepackage[pagebackref,breaklinks,colorlinks,allcolors=iccvblue]{hyperref}

\def\paperID{14} 
\def\confName{ICCV}
\def\confYear{2025}

\newcommand{\YSR}[1]{\textcolor{blue}{#1}}

\title{
\includegraphics[height=20pt]{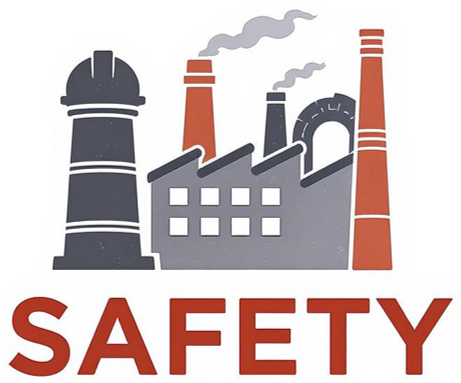} iSafetyBench: A video-language benchmark for safety in industrial environment
}

\author{Raiyaan Abdullah\\
{\tt\small raiyaanabdullah@gmail.com}
\and
Yogesh Singh Rawat\\
{\tt\small yogesh@crcv.ucf.edu}
\and
Shruti Vyas\\
{\tt\small shruti@ucf.edu}
\and
\small University of Central Florida\\
\small Project page: \href{https://isafetybench.github.io}{https://isafetybench.github.io}
}

\begin{document}
\twocolumn[{
\maketitle
\begin{center}
\centering
\captionsetup{type=figure}
\includegraphics[width=1\textwidth]{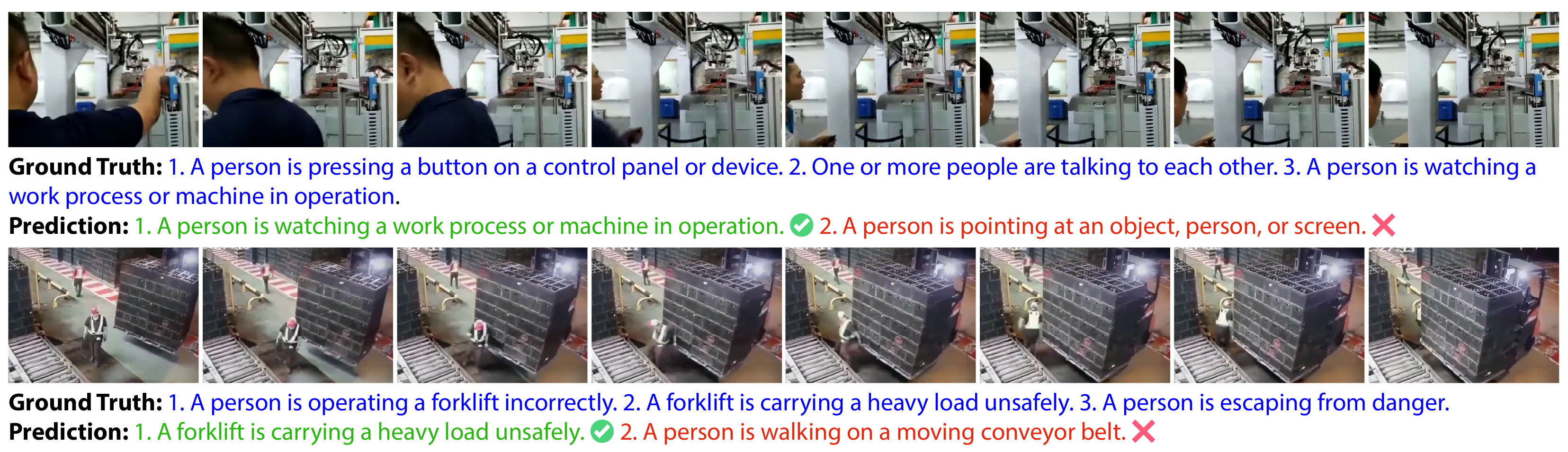}
\caption{\textbf{\textit{Sample multi‐label examples:}} Few frames extracted from videos. Even top performing models such as Ovis2-8B struggle to predict nuanced actions in industrial settings, both in normal (top) and hazardous (bottom) scenarios, often confusing them with visually similar distractors. 
}
\label{fig:teaser}
\end{center}
}]

\begin{abstract}

Recent advances in vision-language models (VLMs) have enabled impressive generalization across diverse video understanding tasks under zero-shot settings. However, their capabilities in high-stakes industrial domains—where recognizing both routine operations and safety-critical anomalies is essential—remain largely underexplored. To address this gap, we introduce \textbf{iSafetyBench}, a new video-language benchmark specifically designed to evaluate model performance in industrial environments across both normal and hazardous scenarios. iSafetyBench comprises \textbf{1,100 video clips} sourced from real-world industrial settings, annotated with \textbf{open-vocabulary, multi-label action tags} spanning \textbf{98 routine} and \textbf{67 hazardous} action categories. Each clip is paired with \textbf{multiple-choice questions} for both \textit{single-label} and \textit{multi-label} evaluation, enabling fine-grained assessment of VLMs in both standard and safety-critical contexts. We evaluate \textbf{eight state-of-the-art video-language models} under zero-shot conditions. Despite their strong performance on existing video benchmarks, these models struggle with iSafetyBench—particularly in recognizing hazardous activities and in multi-label scenarios. Our results reveal significant performance gaps, underscoring the need for more robust, safety-aware multimodal models for industrial applications. iSafetyBench provides a first-of-its-kind testbed to drive progress in this direction. The dataset is available at: \href{https://github.com/iSafetyBench/data}{https://github.com/iSafetyBench/data}.\end{abstract}   
\section{Introduction}
\label{sec:intro}


Recent breakthroughs in multimodal foundation models have brought vision-language models (VLMs) to the forefront of video understanding research. These models, powered by large-scale pretraining and aligned multimodal embeddings, have demonstrated impressive zero-shot capabilities across a range of tasks—including action recognition, video question answering, and temporal localization—on benchmarks such as Kinetics~\cite{kinetics700}, Something-Something~\cite{somethingsomething}, and Ego4D~\cite{ego4d}. As a result, VLMs are rapidly emerging as general-purpose perception engines, capable of interpreting complex real-world activities with minimal task-specific supervision. However, despite their remarkable success on open-domain video datasets, the effectiveness of VLMs in specialized, high-risk domains such as industrial environments remains largely unexplored.

This is a critical gap. Industrial and safety-critical environments present unique challenges for visual understanding: complex machinery, unpredictable human-object interactions, crowded or cluttered scenes, and, most importantly, the need to reliably identify rare but dangerous events. As industries adopt intelligent monitoring systems to reduce risk and improve workplace safety, there is a growing demand for models that can reason about both routine operations and hazardous anomalies. Unfortunately, most existing video benchmarks are not designed with these needs in mind. General-purpose datasets lack safety-specific labels, while industrial or hazard-focused datasets are limited in scope, size, or diversity.

Several prior efforts have tackled action recognition in security and industrial domains~\cite{inhard,timo,ucfcrime,safeunsafe,constructionmeta}, yet these datasets suffer from three key limitations. First, many cover only a narrow set of activities, often choreographed or constrained to specific sites. Second, most are unimodal and closed-set, offering limited flexibility for open-vocabulary or multi-label evaluation—capabilities essential for robust real-world deployment. Third, existing benchmarks typically focus either on routine behaviors or on accidents in isolation, preventing comprehensive evaluation across the full spectrum of operational and hazardous conditions. A benchmark that jointly addresses these limitations—capturing both the complexity of real-world industrial scenes and the subtlety of safety-critical actions—is still missing.

To address these challenges, we introduce \textbf{iSafetyBench}, a new benchmark for open-vocabulary video-language evaluation in industrial and safety-sensitive scenarios. The dataset consists of \textbf{1,100 real-world video clips} curated from publicly available online sources, covering diverse locations such as factories, warehouses, construction sites, parking lots, and retail spaces. It is organized into two balanced splits: \textit{normal routine actions} (e.g., assembling parts, operating machinery, inspecting equipment) and \textit{dangerous hazardous events} (e.g., structural collapses, fires, manual handling injuries, vehicle accidents). Each video is annotated with multiple open-vocabulary action labels and paired with carefully crafted \textbf{multiple-choice questions} for both \textit{single-label} and \textit{multi-label} evaluation.

We evaluate \textbf{eight state-of-the-art video-language models}, including both open-source and closed-source systems, under a challenging zero-shot setting. Despite strong performance on conventional benchmarks, these models show notable weaknesses on iSafetyBench—particularly in identifying hazardous events and reasoning over multiple simultaneous actions. For instance, average model accuracy drops significantly when moving from routine to hazardous clips, and from multi-label to single-label questions. These results reveal critical gaps in current VLM capabilities and motivate future research in safety-aware multimodal understanding.

Our key contributions are:
\begin{itemize}
\item We introduce \textbf{iSafetyBench}, a new open-vocabulary, multi-label video-language benchmark focused on industrial and safety-critical actions. It consists of a diverse dataset of \textbf{1,100 real-world video clips} with detailed annotations across \textbf{98 routine} and \textbf{67 hazard} action categories, supporting both \textit{single} and \textit{multi-label} MCQ-based evaluation.
\item We benchmark \textbf{eight recent vision-language models} under zero-shot settings and provide a comprehensive analysis of their performance on routine vs. hazardous actions and single vs. multi-label setups.
\item We highlight key limitations of current VLMs in recognizing safety-critical events and provide a challenging new testbed to drive progress in this direction.
\end{itemize}
\begin{table*}[t!]
\centering
\resizebox{17.5cm}{!}{
\begin{tabular}{lcccccccccc}
\toprule
Dataset 
  & \makecell{Normal\\scenarios}
  & \makecell{Dangerous\\scenarios}
  & Multi-label 
  & \makecell{Textual\\data}
  & \makecell{Environment\\type(s)} 
  & \makecell{Set\\type} 
  & \makecell{\# Normal\\actions} 
  & \makecell{\# Non-critical\\anomaly actions} 
  & \makecell{\# Danger/hazard\\actions} 
  & \makecell{\# High-level\\categories} \\
\midrule
UCF-Crime \cite{ucfcrime}                        & \cmark & \cmark & \xmark & \xmark & Multiple & Closed & 0 & 0 & 13 & 0 \\
InHARD \cite{inhard}                            & \cmark & \xmark & \xmark & \xmark & Single & Closed & 74 & 0 & 0 & 14 \\
TIMo \cite{timo}                                & \cmark & \xmark & \xmark & \xmark & Single & Closed & 35 & 21 & 0 & 20\footnotemark \\
OpenPack \cite{openpack}                        & \cmark & \cmark & \xmark & \xmark & Single & Closed & 43 & 44 & 1 & 17 \\
Safe/Unsafe Behaviours \cite{safeunsafe}        & \cmark & \cmark & \xmark & \xmark & Single & Closed & 4 & 0 & 4 & 2 \\
Construction Meta Action \cite{constructionmeta} & \cmark & \cmark & \xmark & \xmark & Single & Closed & 1 & 0 & 6 & 0 \\
\midrule
\textbf{iSafetyBench(Ours)}                          & \cmark & \cmark & \cmark & \cmark & Multiple & Open & 98 & 0 & 67 & 18 \\
\bottomrule
\end{tabular}
}

\caption{\textbf{\textit{Dataset Comparison:}} Comparison with existing security and industrial datasets. Our proposed iSafetyBench is the only open‐vocabulary (open-set), multi‐label benchmark that pairs textual questions with clips. It covers multiple environment types with a high number of normal and danger/hazard actions in 18 high-level categories.  
}
\label{tab:comparison_industrial}
\end{table*}

\section{Related Work}
\label{sec:relatedwork}

\paragraph{General Video Action Recognition.}
Large-scale video action recognition datasets have played a key role in driving progress in video understanding. Datasets such as Kinetics-700\cite{kinetics700}, ActivityNet\cite{activitynet}, Something-Something\cite{somethingsomething}, Charades\cite{charades}, EPIC-KITCHENS\cite{epickitchens}, and Ego4D\cite{ego4d} offer broad coverage of human activities across everyday contexts. These benchmarks have been widely used to train and evaluate both unimodal and multimodal models. While valuable for assessing generic video understanding, these datasets lack annotations for high-risk or safety-critical scenarios, and they do not target the unique characteristics of industrial environments such as machinery interaction, equipment handling, or dangerous incidents.

\paragraph{Surveillance and Anomaly Detection.}
Several datasets focus on video security and anomaly detection, often using CCTV or drone footage in public spaces. Representative examples include UCF-Crime\cite{ucfcrime}, ShanghaiTech Campus\cite{shanghaitech}, XD-Violence\cite{xdviolence}, UBnormal\cite{ubnormal}, and MEVA~\cite{meva}. These datasets typically capture abnormal behaviors (e.g., fights, thefts, running in restricted zones) and are used for binary anomaly classification or temporal localization. However, most of these benchmarks are either limited to a small set of coarse event types or lack the open-vocabulary, multi-label setups needed for fine-grained evaluation. Furthermore, they rarely capture industrial hazards such as equipment failure, falling objects, or operational accidents.

\begin{figure}[t!]
    \centering
    \includegraphics[width=\linewidth]{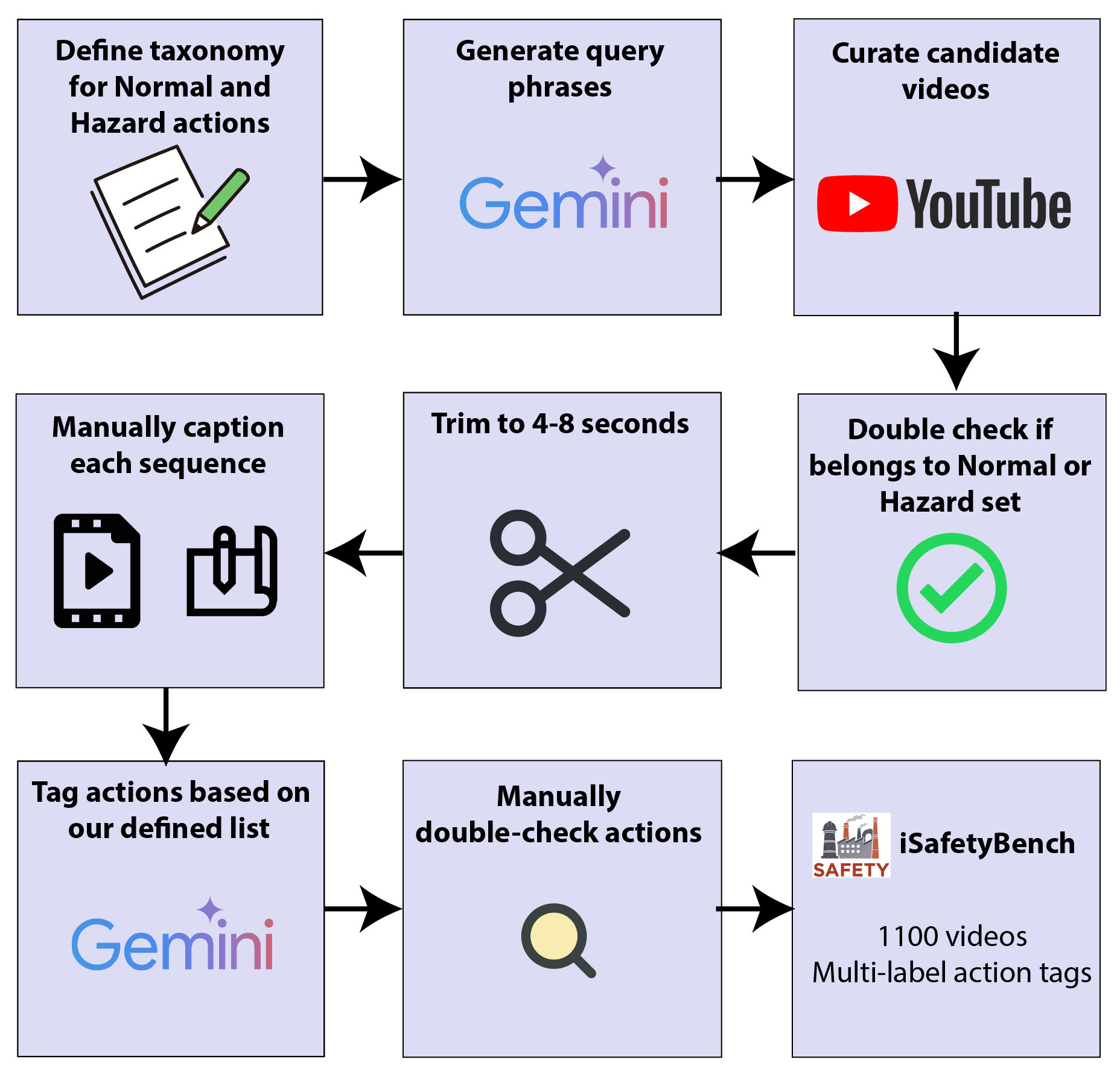}
    \caption{\textbf{\textit{Overview of iSafetyBench Curation:}} Outline showing dataset generation pipeline used to curate iSafetyBench.
    }
    \label{fig:flow}
\end{figure}

\begin{figure*}[t!]
    \centering
    \includegraphics[width=\linewidth]{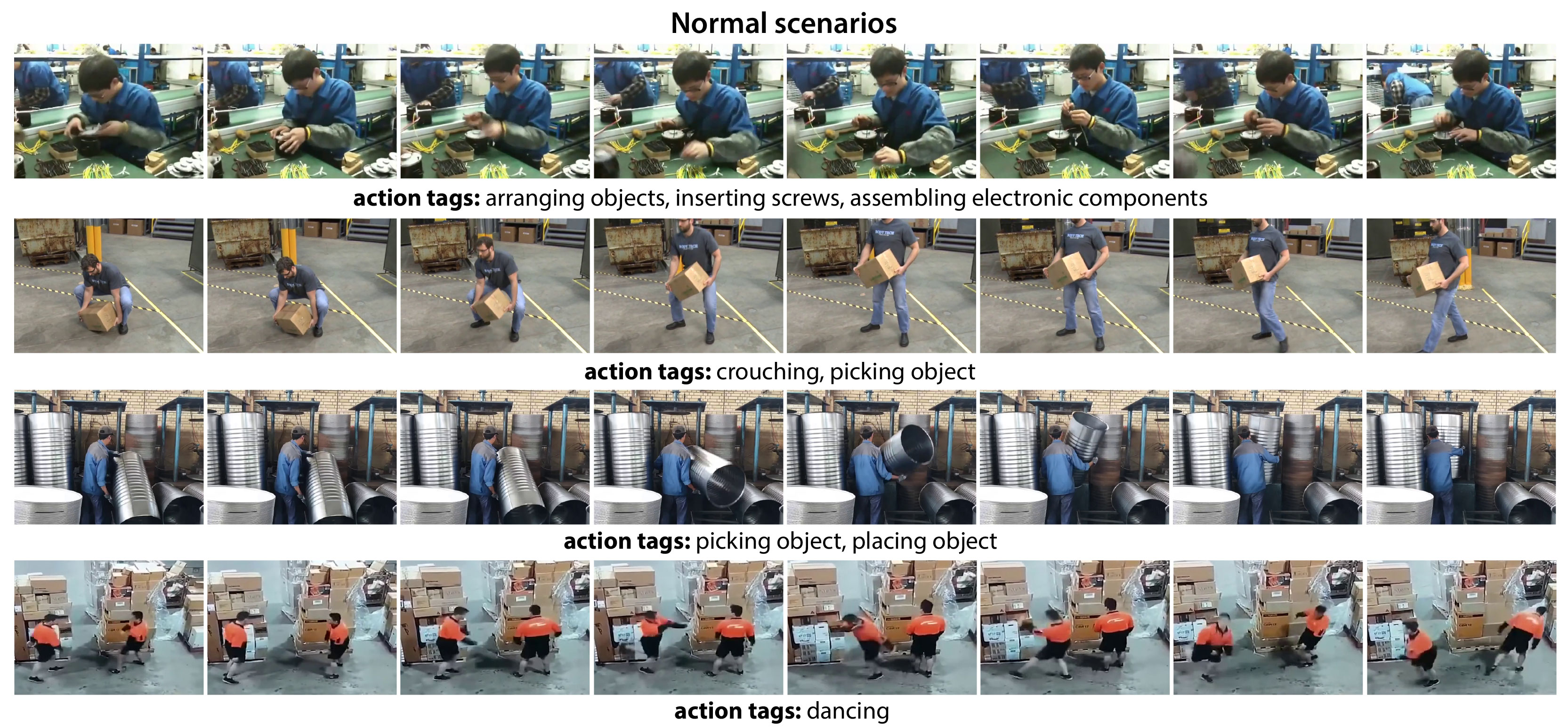}
    \caption{\textbf{\textit{Examples from the normal dataset:}} Our dataset encompasses a wide spectrum of actions, from routine industrial tasks such as loading and inspecting to unexpected behaviors like dancing in a warehouse.
    }

    \label{fig:normal_dataset_sample}
\end{figure*}

\paragraph{Industrial and Safety-Focused Datasets.}
A smaller set of benchmarks specifically targets industrial or safety-related tasks. InHARD\cite{inhard} focuses on human action recognition in industrial settings but lacks coverage of dangerous or anomalous events. TIMo\cite{timo} captures manufacturing tasks but is restricted to indoor, scripted setups. OpenPack\cite{openpack} focuses on package handling activities with limited hazard coverage. Datasets like Safe/Unsafe Behaviors\cite{safeunsafe} and Construction Meta Action~\cite{constructionmeta} include accident events or unsafe actions, but typically involve very few categories, limited video volume, and are often staged. Overall, these datasets are either narrow in scope, lack routine action coverage, or are closed-set with predefined labels—limiting their use for evaluating modern vision-language models in open-world safety-critical applications.

\footnotetext{InHARD authors do not explicitly provide any category, we formed it based on the actions.}

\paragraph{Multimodal and Open-Vocabulary Benchmarks.}
Recent interest in multimodal and vision-language models has led to the creation of benchmarks that test video-language alignment, temporal grounding, and compositional reasoning. Datasets such as VidSitu\cite{vidsitu}, Action Genome\cite{actiongenome}, EgoSchema\cite{egoschema}, and VideoChatBench\cite{videollama} emphasize reasoning over text-video pairs, but none focus on the unique demands of industrial or safety settings. Notably, these benchmarks are designed for rich linguistic understanding but do not cover rare, high-risk events or multi-label classification. Similarly, new long-video benchmarks like LVBench\cite{lvbench}, LongVideoBench\cite{longvideobench}, and TempCompass~\cite{tempcompass} test temporal reasoning but again lack focus on safety or hazard detection.

In summary, while prior work has made substantial progress in both action recognition and video-language modeling, there remains a significant gap in benchmarking VLMs for industrial and safety-critical applications. Existing datasets either lack diversity, scale, or multimodal evaluation protocols needed to assess zero-shot generalization in real-world risk-sensitive environments. \textbf{iSafetyBench} fills this gap by providing a large, diverse, and open-vocabulary benchmark designed to jointly evaluate models on both routine industrial activities and rare hazardous events under realistic multi-label and zero-shot settings.

\section{iSafetyBench: Benchmark Details}
\label{sec:benchmark}


We introduce \textbf{iSafetyBench}, a new video benchmark designed to evaluate the capabilities of vision-language models (VLMs) in industrial environments. Unlike existing benchmarks, which primarily focus on general activities or staged scenarios, iSafetyBench captures the complexity of real-world industrial setups across both routine operations and hazardous incidents. The dataset comprises 1,100 short video clips (4–8 seconds each), collected from diverse environments such as factories, warehouses, construction sites, and retail spaces. Each clip is annotated with multiple open-vocabulary action labels, and paired with multiple-choice questions (MCQs) to facilitate structured evaluation. We support both single-correct and multiple-correct answer settings, enabling robust assessment of discriminative and inclusive model capabilities. Importantly, the benchmark is designed for zero-shot evaluation, without any fine-tuning or task-specific adaptation, to assess generalization in safety-critical domains.

\begin{figure*}[t!]
    \centering
    \includegraphics[width=\linewidth]{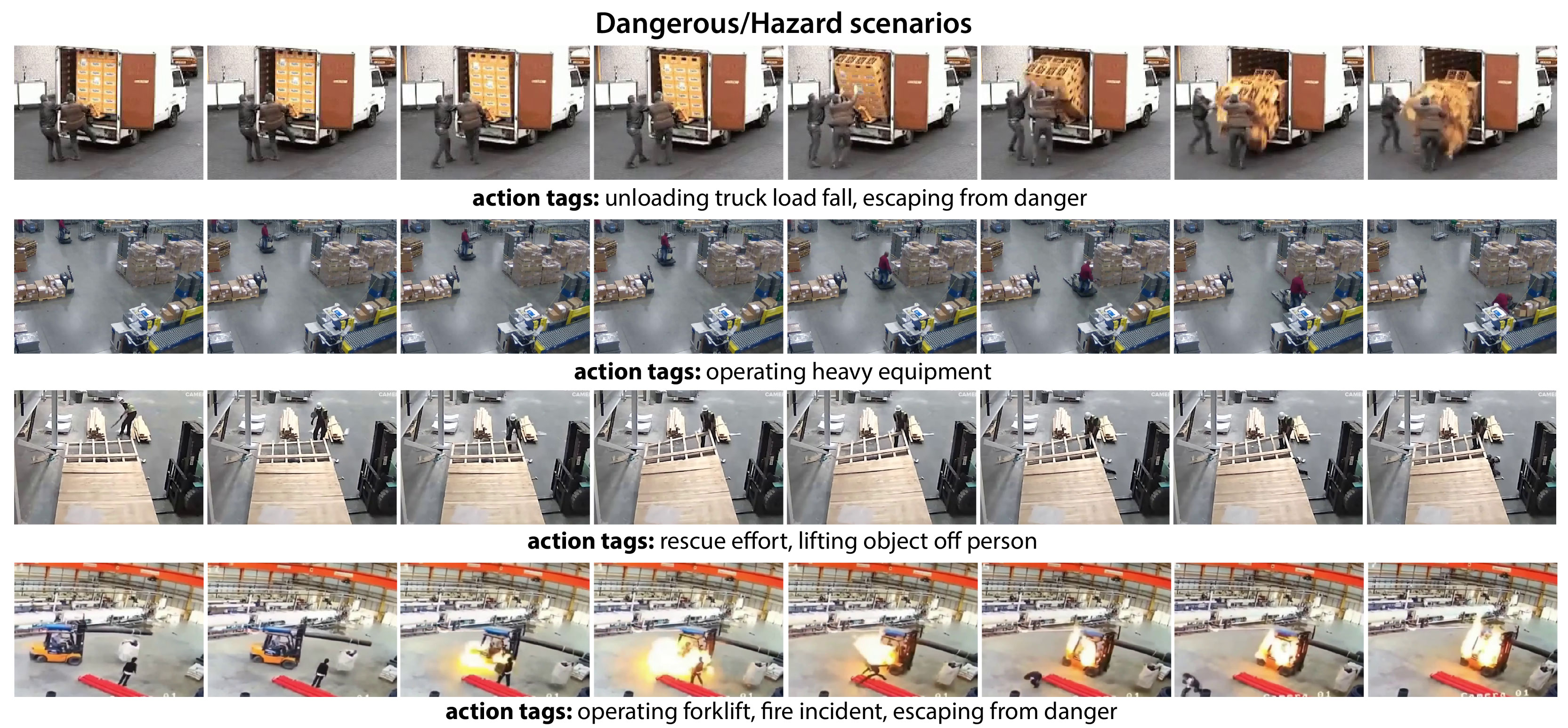}
    \caption{\textbf{\textit{Examples from the danger/hazard dataset:}} Our dataset includes various hazardous scenarios, ranging from load collapses and personal injuries to emergency incidents like fire.
    }
    \label{fig:accident_dataset_sample}
\end{figure*}

\begin{figure*}[t!]
  \centering
  \begin{subfigure}{0.49\linewidth}
    \includegraphics[width=1\linewidth]{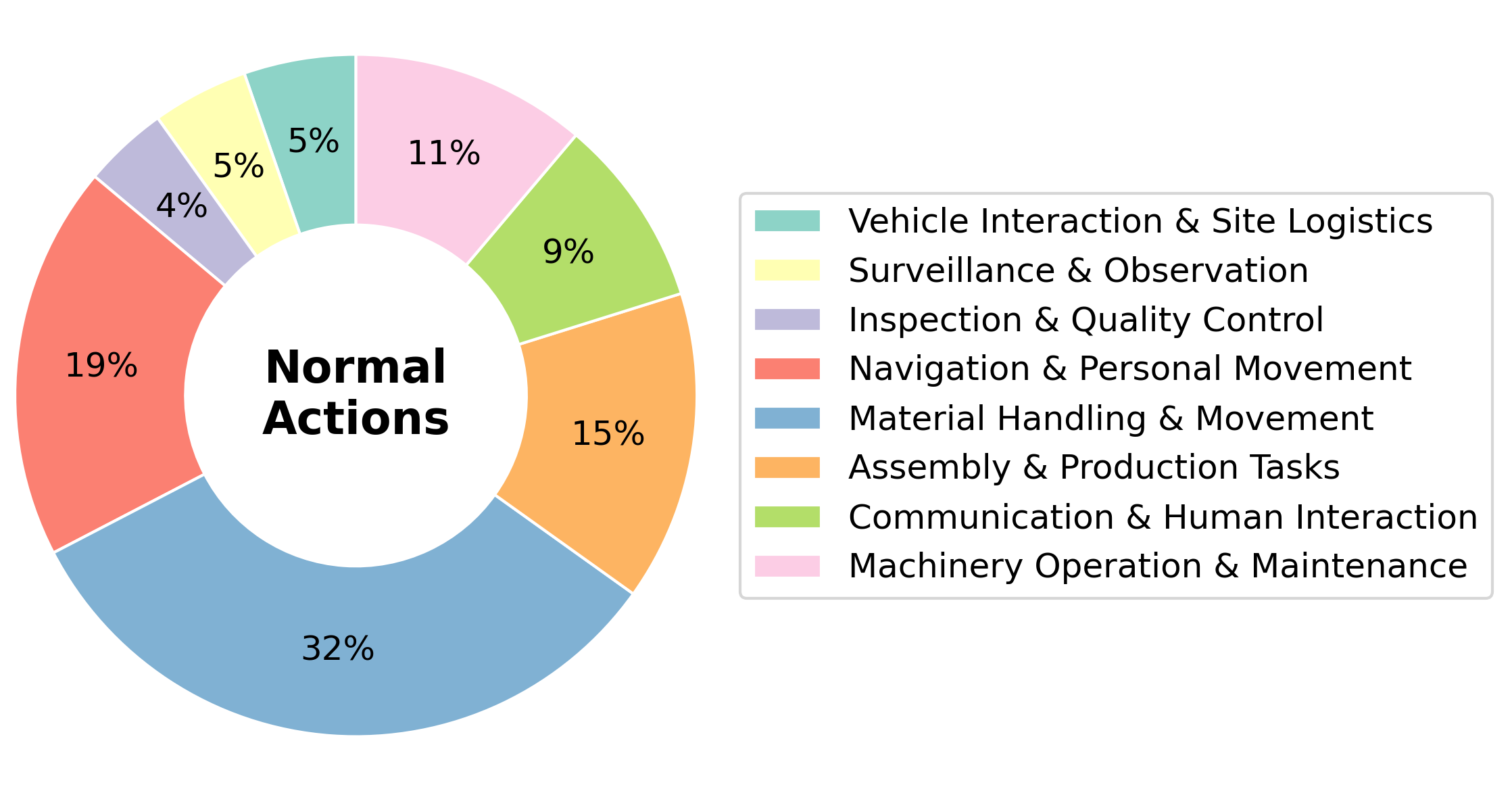}
  \end{subfigure}
  \hfill
  \begin{subfigure}{0.49\linewidth}
    \includegraphics[width=1\linewidth]{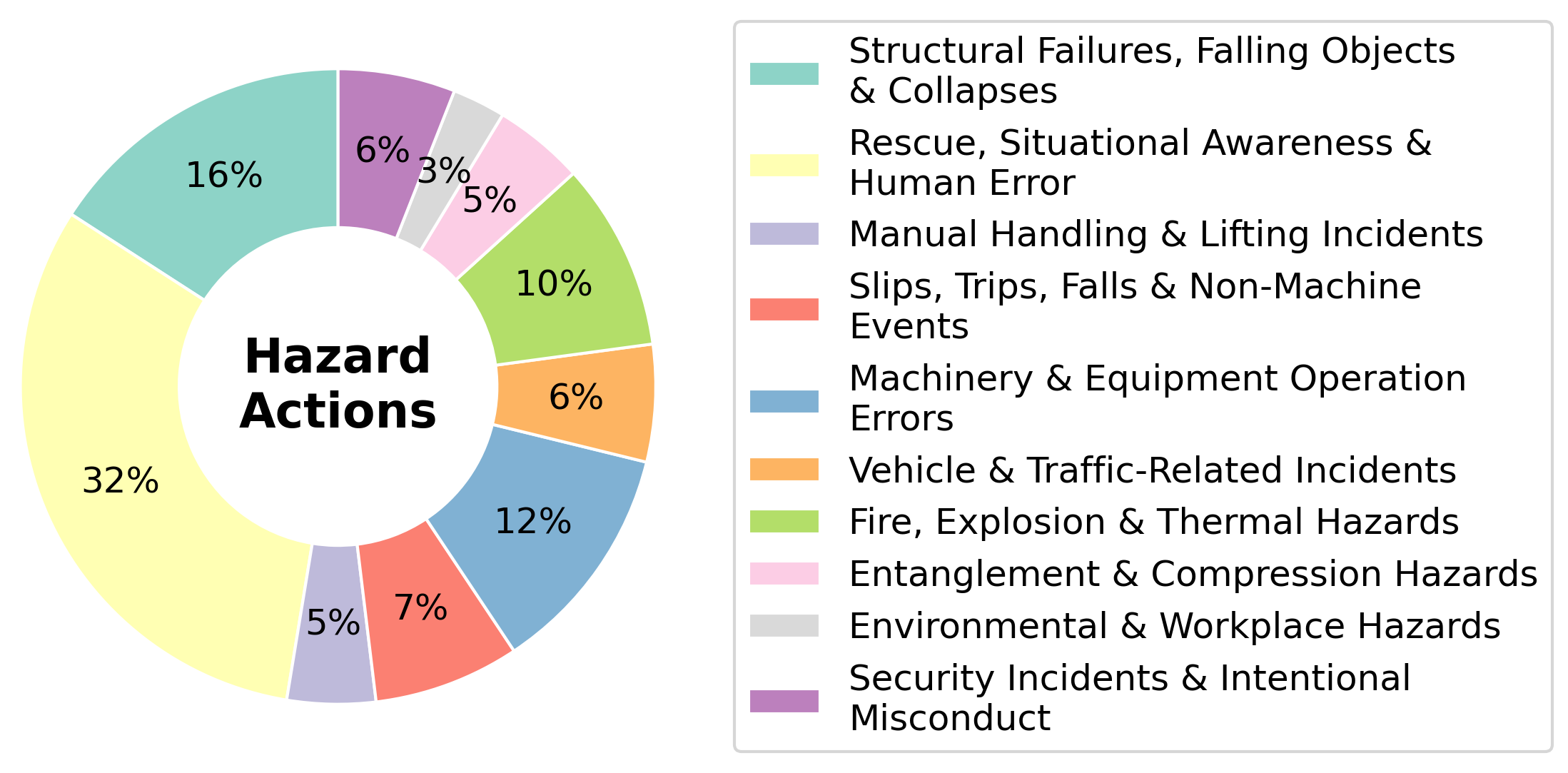}
  \end{subfigure}
  \caption{\textbf{\textit{Dataset distribution:}} Distribution of action tag categories in \textbf{iSafetyBench}.
  }
  \label{fig:category_distribution}
\end{figure*}

\subsection{Dataset Curation Process}

The construction of iSafetyBench followed a multi-stage pipeline focused on coverage, action diversity, and annotation quality. We began by defining a dual taxonomy of actions relevant to industrial settings, dividing them into routine (normal) actions and safety-critical (hazardous) events. This taxonomy was informed by workplace operation manuals, industrial safety guidelines, and publicly available accident reports. We then used Gemini 2.5 Pro to generate keyword-based search queries for each action tag, which were used to retrieve candidate videos from YouTube. All videos were manually reviewed to ensure relevance and trimmed into 4–8 second clips centered around the target action. This preprocessing step minimizes contextual distractions and ensures clarity of action content. An overview of the whole process is given in \cref{fig:flow}

\paragraph{Normal Actions.}
For normal industrial behavior, we define 98 open-vocabulary action labels grouped into eight functional categories: Material Handling \& Movement (e.g., pouring, lifting), Assembly \& Production Tasks (e.g., screwing, sealing boxes), Machinery Operation \& Maintenance (e.g., operating drills or presses), Vehicle Interaction \& Site Logistics (e.g., loading/unloading trucks), Inspection \& Quality Control (e.g., measuring, photographing), Communication \& Human Interaction (e.g., assisting, pointing), Navigation \& Personal Movement (e.g., walking, putting on PPE), and Surveillance \& Observation (e.g., monitoring via camera or flashlight). These categories reflect the breadth of activities typical to real-world industrial workflows.

\paragraph{Hazardous Actions.}
To capture the wide range of dangerous scenarios, we define 67 hazard labels across ten categories: Machinery \& Equipment Operation Errors (e.g., unsafe forklift usage), Entanglement \& Compression Hazards (e.g., clothing caught in machines), Structural Failures and Collapses (e.g., shelf or wall collapse), Fire, Explosion \& Thermal Hazards (e.g., fire outbreak, equipment overheating), Manual Handling \& Lifting Incidents (e.g., lifting injuries, strain-related accidents), Vehicle \& Traffic Related Incidents (e.g., collisions, persons hit by vehicles), Rescue \& Situational Awareness (e.g., evacuations, aiding injured workers), Environmental Hazards (e.g., unsafe floors, sparks), Slips, Trips \& Falls (e.g., falls from height, tripping), and Security \& Misconduct (e.g., altercations, vandalism). This taxonomy is grounded in real accident case studies and industrial safety documentation.

\subsection{Video Retrieval and Preprocessing}
To collect video data, we used Gemini 2.5 Pro to generate diverse query phrases tailored to each action label in both the normal and hazard taxonomies. These queries were used to search YouTube, from which we curated candidate videos. Annotators then manually reviewed the clips and identified segments clearly depicting a single or small group of relevant actions. Each selected segment was trimmed to 4–8 seconds in length to ensure focus and clarity. This manual curation process resulted in high-quality, action-centric video clips that reflect real-world variability in lighting, camera angle, and background clutter.

\subsection{Annotation and Review}
Each clip was annotated using a semi-automated pipeline. A human annotator first wrote a free-form caption describing all observable actions involving humans, tools, or vehicles. These captions were processed through Gemini 2.5 Pro, which proposed a list of potential action tags from our predefined taxonomy. The annotator then reviewed and refined these tags to ensure semantic consistency and coverage. This process balances scalability and precision, ensuring high-quality annotations while maintaining efficient throughput across the large-scale dataset.

\begin{table*}[t!]
\centering
\resizebox{17.5cm}{!}{
\begin{tabular}{l
                c c c c
                c c c c
                c c}
\toprule
 & \multicolumn{4}{c}{\textbf{Normal}} 
 & \multicolumn{4}{c}{\textbf{Danger/Hazard}} 
 & \multicolumn{2}{c}{\textbf{Average of Both}} \\
\cmidrule(lr){2-5} \cmidrule(lr){6-9} \cmidrule(l){10-11}
\textbf{Model} 
 & \textbf{Single} 
 & \multicolumn{3}{c}{\textbf{Multi}} 
 & \textbf{Single} 
 & \multicolumn{3}{c}{\textbf{Multi}} 
 & \textbf{Single} 
 & \textbf{Multi} \\
\cmidrule(lr){2-2} \cmidrule(lr){3-5} \cmidrule(lr){6-6} \cmidrule(lr){7-9} \cmidrule(lr){10-10} \cmidrule(l){11-11}
 & \textbf{Acc\,(\%)} 
 & \textbf{Precision} 
 & \textbf{Recall} 
 & \textbf{F1 Score} 
 & \textbf{Acc\,(\%)} 
 & \textbf{Precision} 
 & \textbf{Recall} 
 & \textbf{F1 Score} 
 & \textbf{Acc\,(\%)} 
 & \textbf{F1 Score} \\
\midrule
\textbf{Ovis2-8B}               & 47.3 & 47.6 & \underline{\textbf{71.3}} & \underline{\textbf{53.4}} & \underline{\textbf{40.3}} & 45.0 & 54.1 & 46.2 & \underline{\textbf{43.8}} & 49.8  \\
\textbf{InternVL2.5-8B-MPO}     & 42.2 & 47.2 & 64.1                     & 50.8                     & 38.3                     & 47.0 & \underline{\textbf{57.7}} & \underline{\textbf{49.0}} & 40.25                    & \underline{\textbf{49.9}} \\
\textbf{Qwen2.5-VL-7B-Instruct} & 46.9 & 44.5 & 62.7                     & 49.2                     & 33.6                     & 40.5 & 47.9                     & 41.7                     & 40.25                    & 45.45 \\
\textbf{VideoLLaMA3-7B}         & 38.9 & 49.6 & 37.4                     & 39.7                     & 32.7                     & 46.2 & 34.1                     & 36.5                     & 35.8                     & 38.1  \\
\textbf{VideoChat-Flash-7B}     & 31.0 & 36.3 & 44.2                     & 33.6                     & 26.8                     & 36.0 & 39.1                     & 31.0                     & 28.9                     & 32.3  \\
\textbf{Oryx-7B}                & 25.0 & 37.8 & 41.7                     & 30.3                     & 21.5                     & 34.7 & 37.2                     & 26.6                     & 23.25                    & 28.45 \\
\textbf{Valley-Eagle-7B}        & \underline{\textbf{48.8}} & \underline{\textbf{59.1}} & 47.7 & 48.7 & 35.9 & \underline{\textbf{54.5}} & 36.8 & 40.9 & 42.35 & 44.8  \\
\textbf{GPT-4o}                 & 40.3 & 50.0 & 59.1                     & 51.6                     & 37.3                     & 49.3 & 45.4                     & 45.1                     & 38.8                     & 48.35 \\
\bottomrule
\end{tabular}
}
\caption{\textbf{\textit{Performance Evaluation:}} Performance of VLMs on normal and danger/hazard scenarios of iSafetyBench.}
\label{tab:model_results}
\end{table*}

\begin{figure*}[t!]
\centering
\includegraphics[width=\linewidth]{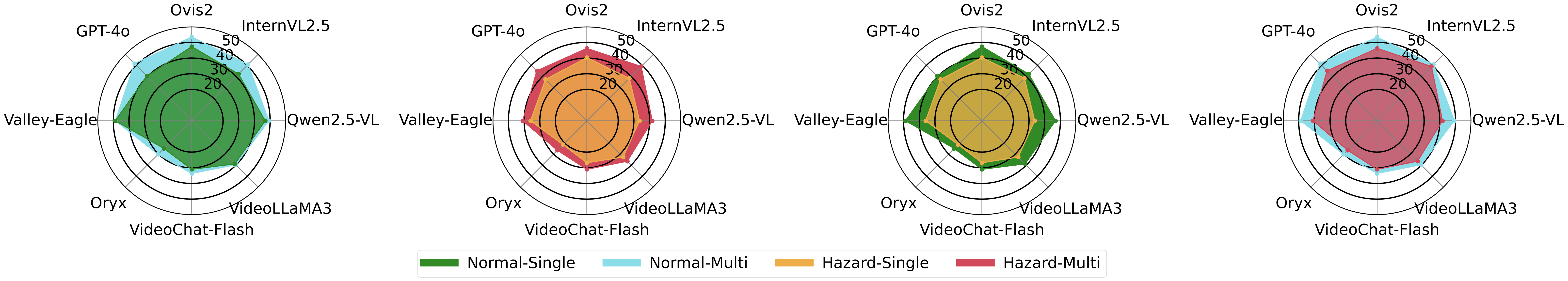}
\caption{
\textbf{\textit{Performance comparison of models:}} We observe a consistent drop in accuracy from multi‐choice to single‐choice evaluations, and from normal routine to hazardous video scenarios.
}
 \label{fig:model_scores}
 \vspace{-10pt}
\end{figure*}

\subsection{Dataset Statistics}
The final dataset comprises 1,100 video clips, with 680 labeled as normal (\cref{fig:normal_dataset_sample}) and 420 labeled as hazard scenarios (\cref{fig:accident_dataset_sample}). Across all videos, we obtain 98 distinct normal action labels and 67 hazard labels, resulting in 1,468 annotated normal action instances and 888 hazard instances. On average, each clip contains 2–3 annotated actions, with substantial multi-label overlap due to the complexity of real-world scenarios. The dataset spans both indoor and outdoor environments, varied weather and lighting conditions, and includes both first-person and third-person viewpoints. \cref{fig:category_distribution} visualizes the frequency distribution of action categories, highlighting the prevalence of routine tasks such as material handling and the prominence of situational awareness and structural failures in hazard scenarios.

\subsection{Evaluation Setup}
To evaluate model performance in a structured setting, we pair each video with multiple-choice questions designed to test understanding of observed actions. For every action-labeled video, we construct two types of MCQs. In the single-correct-choice setting, the question targets one ground-truth action and includes 15 distractors, with only one correct answer. In the multiple-correct-choice setting, several actions may be valid, and the model must identify all applicable answers from a list of 16 options. This dual formulation supports both precision-oriented and recall-oriented evaluation.

Distractor options are generated using Gemini 2.5 Pro by selecting actions that are semantically or visually similar to the ground truth but incorrect. These options are manually verified and refined to ensure that each question is challenging and unambiguous, avoiding trivial elimination.

\paragraph{Evaluation Metrics.}
We adopt accuracy for single-correct-choice questions, where a response is correct only if the model selects the single ground-truth action. For multiple-correct-choice questions, we compute precision, recall, and F1 score based on the set of selected versus true labels. Precision captures how many of the predicted actions are correct, recall captures how many of the ground-truth actions are retrieved, and F1 score balances the two. This dual-metric setup allows us to evaluate both fine-grained decision-making and broader scene understanding under open-vocabulary, zero-shot settings.

\begin{figure*}[t!]
    \centering
    \includegraphics[width=\linewidth]{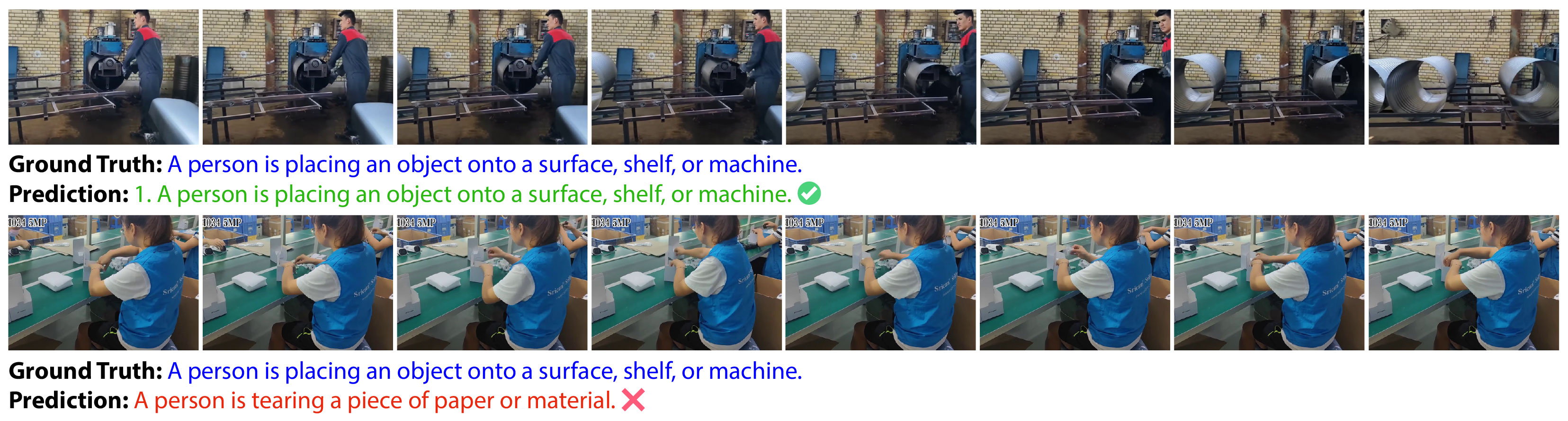}
    \caption{\textbf{\textit{Success and failure examples for normal single-choice MCQs:}} In the top sequence, Ovis2-8B correctly sees someone putting something down on a surface. In the bottom sequence, it mistakes that motion for tearing paper.
    }
    \label{fig:normal_dataset_sample_eval}
\end{figure*}

\begin{figure*}[t!]
    \centering
    \includegraphics[width=\linewidth]{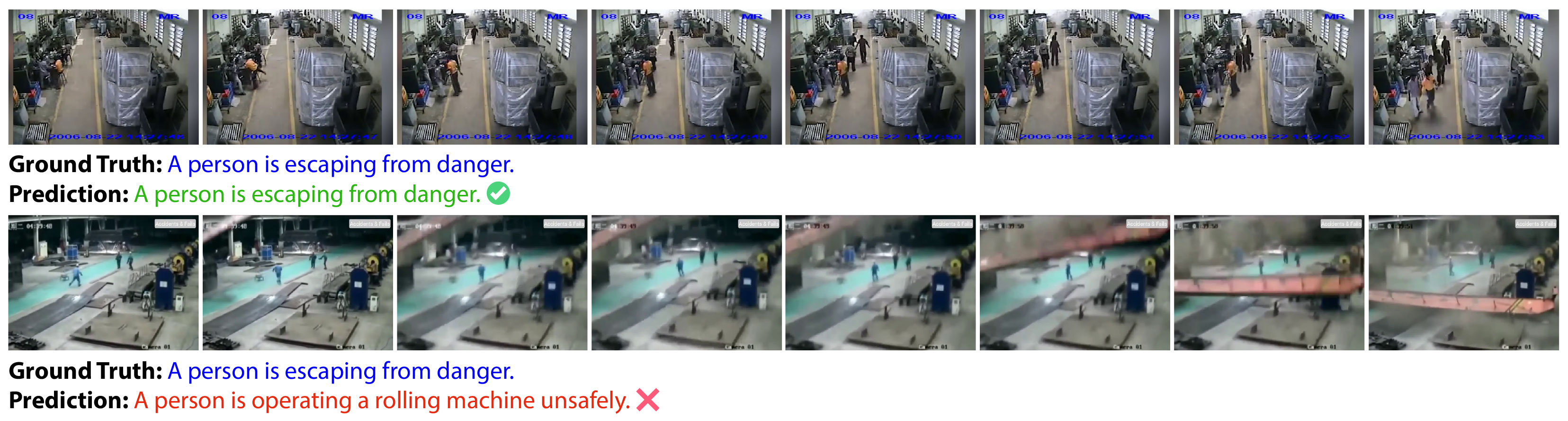}
    \caption{\textbf{\textit{Success and failure examples for hazard single-choice MCQs:}} In the top sequence, Ovis2-8B correctly spots someone moving away from danger. In the bottom sequence, it mistakes that action for someone running a rolling machine unsafely.
    }
    \label{fig:hazard_dataset_sample_eval}
\end{figure*}

\section{Experiments and Results}

\paragraph{Video Language Models:} We evaluate eight state-of-the-art Vision-Language Models (VLMs) on our normal and accident video datasets. The open-source models include Ovis2-8B \cite{ovis}, InternVL2.5-8B-MPO \cite{internvl}, Qwen2.5-VL-7B-Instruct \cite{qwen25vl}, VideoLLaMA3-7B \cite{videollama}, VideoChat-Flash-7B \cite{videochatflash}, Valley-Eagle-7B \cite{valley}, and Oryx-7B \cite{oryx}, with parameter counts ranging from 7 to 8 billion. GPT-4o \cite{gpt4o} is the only closed-source model evaluated in our study. We evaluate all the videos in our dataset in a zero-shot setting.

\subsection{Results and analysis}
\begin{figure*}[t!]
    \centering
    \includegraphics[width=\linewidth]{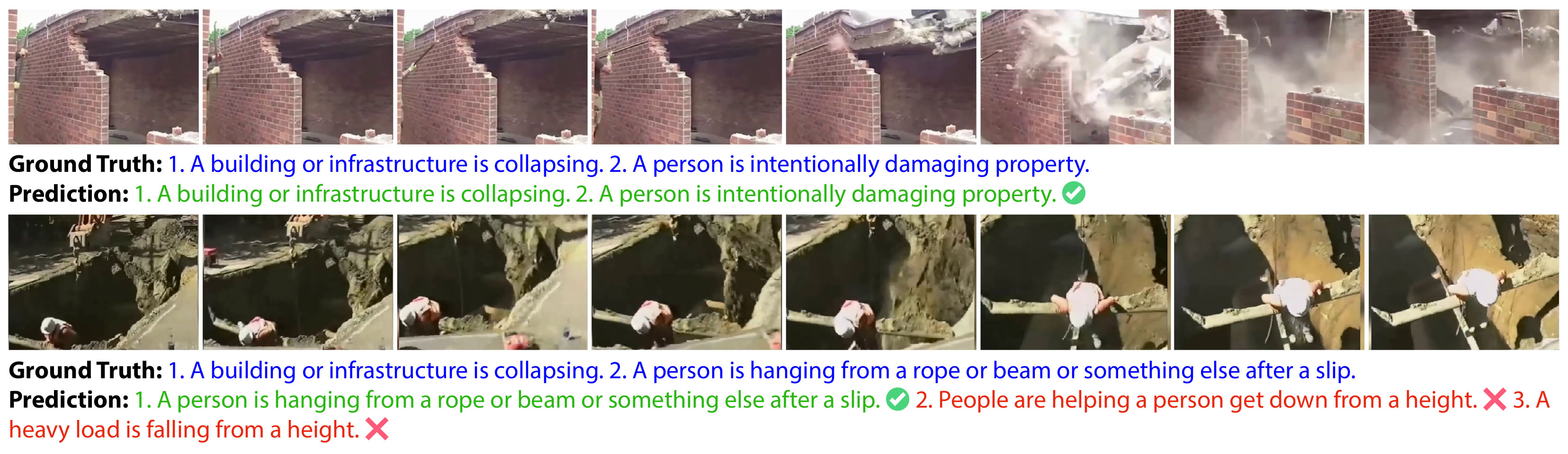}
    \caption{\textbf{\textit{Success and failure examples for hazard multi-choice MCQs:}}  In the top sequence, Ovis2-8B correctly picks out both the building collapsing and someone tearing down the wall. In the bottom sequence, it sees the person hanging after a slip but then wrongly labels the scene as people helping them down and a heavy load falling.
    }
    \label{fig:hazard_dataset_sample_eval_multi}
\end{figure*}

As seen in \cref{tab:model_results} and \cref{fig:model_scores}, most models score between 35\% and 50\%. This shows that their overall understanding of industrial and security actions remains limited. The highest result is 53.4\% by Ovis2-8B on normal multiple-correct-choice videos. This emphasizes that even the most advanced models struggle to interpret these scenarios accurately. 
~\cref{fig:normal_dataset_sample_eval}, ~\cref{fig:hazard_dataset_sample_eval}, and ~\cref{fig:hazard_dataset_sample_eval_multi} present qualitative results from \textbf{iSafetyBench}, showcasing representative examples from both normal and hazardous scenarios under single-label and multi-label evaluation settings. These figures highlight the model outputs from Ovis2-8B, one of the best-performing vision-language models in our study. For each setting, we illustrate both successful predictions—where the model correctly identifies all relevant actions—and failure cases, where the model misinterprets the scene or misses key safety-critical cues. 
\paragraph{Single‐label vs. Multi‐label Performance:} Across our eight models, the average jump from single‐label accuracy to multi‐label F1 is 4.6\% for normal actions and 6.3\% for hazard actions. The largest individual improvement on normal actions is 11.3\% (GPT-4o: 40.3\% → 51.6\%), and on hazard actions it is 10.7\% (InternVL2.5-8B-MPO: 38.3\% → 49.0\%). This consistent uplift underscores the benefit of multi‐label evaluation in granting partial credit for correctly identified subsets of actions. In the \emph{Average of Both} columns of \cref{tab:model_results}, it is seen multi scores exceed single‐label accuracy for most models. InternVL2.5-8B-MPO exhibits the largest mean uplift (40.25\% vs. 49.90\%), followed by GPT-4o (38.80\% vs. 48.35\%) and Ovis2-8B (43.80\% vs. 49.80\%). Only Valley-Eagle-7B shows a smaller gap (42.35\% vs. 44.80\%).
\paragraph{Normal vs. Hazard Performance:} On average, models score higher on normal‐action questions than on hazard questions. Across the eight models, the mean single‐label accuracy drops an average gap of 6.7\%. In the multi‐label setting, the average gap is 5\%. The largest drop in single‐label performance occurs for Qwen2.5-VL-7B-Instruct (46.9\% → 33.6\%, a 13.3\% gap), while Valley-Eagle-7B exhibits the biggest multi‐label decline (48.7\% → 40.9\%, a 7.8\% gap). These results show that recognizing rare or safety‐critical events remains substantially more challenging than routine activities.

\begin{figure*}[ht!]
  \centering
  \begin{subfigure}{0.49\linewidth}
    \includegraphics[width=1\linewidth]{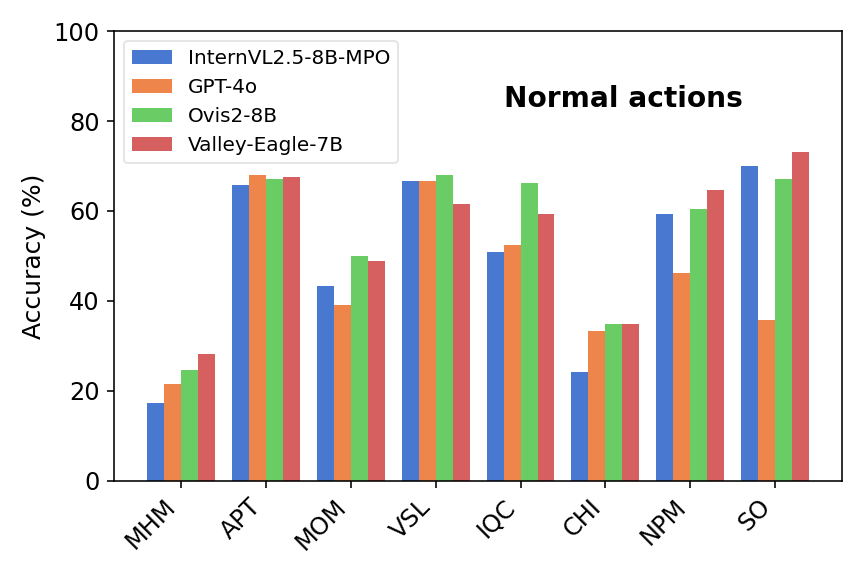}
  \end{subfigure}
  \hfill
  \begin{subfigure}{0.49\linewidth}
    \includegraphics[width=1\linewidth]{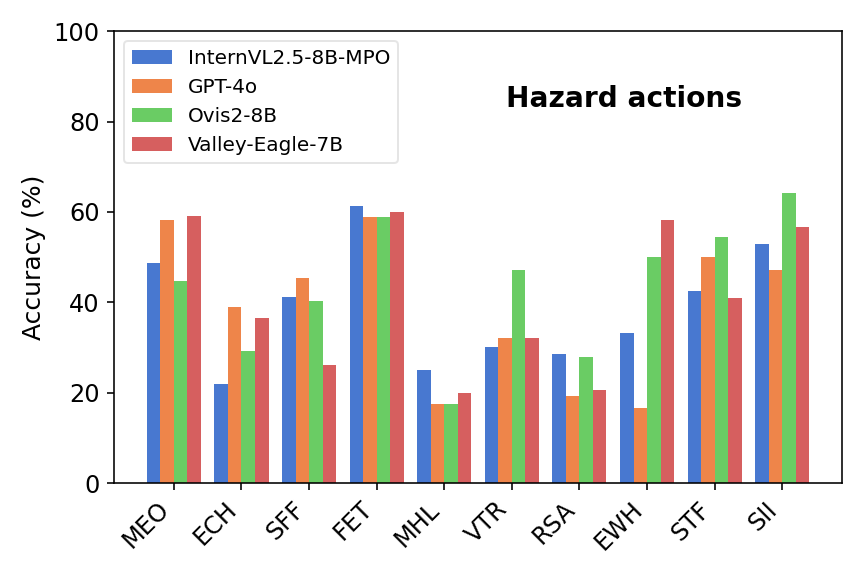}
  \end{subfigure}
  \vspace{-5pt}
  \caption{\textbf{\textit{Performance Analysis Across Categories:}} Accuracy distribution across action categories for the four top‐performing models. The models show lower performance in categories such as Material Handling \& Movement (MHM), Communication \& Human Interaction (CHI), Rescue, Situational Awareness \& Human Error (RSA), and Manual Handling \& Lifting Incidents (MHL).
  }
  \label{fig:acc_distribution}
\end{figure*}

\paragraph{Category‐wise Performance:}
We examined the category‐wise performance of our four top‐performing models \cite{ovis,qwen25vl,gpt4o,valley}. As shown in \cref{fig:acc_distribution}, all four models achieve their highest accuracy on structured, object‐centric normal actions such as Assembly \& Production Tasks (APT) and Vehicle Interaction \& Site Logistics (VSL) with scores clustered around 65–70\%, followed closely by Surveillance \& Observation (SO). In contrast, Material Handling \& Movement (MHM) and Communication \& Human Interaction (CHI) remain challenging (20–35\%). For hazard actions, events with strong distinctive visual cues like Fire, Explosion \& Thermal Hazards (FET) and Security Incidents \& Intentional Misconduct (SII)—reach 55–65\% accuracy, while subtle, context‐dependent hazard categories such as Rescue, Situational Awareness \& Human Error (RSA) and Manual Handling \& Lifting Incidents (MHL) fall below 30\%.

\section{Conclusion}
\label{sec:conclusion}


In this work, we introduced \textbf{iSafetyBench}, a new benchmark designed to assess the capabilities of vision-language models (VLMs) in industrial and safety-critical scenarios. Unlike prior datasets that focus on generic or narrowly-defined tasks, iSafetyBench covers a broad spectrum of real-world activities across both routine operations and hazardous incidents. The dataset supports open-vocabulary multi-label annotation and structured multiple-choice question answering, enabling zero-shot evaluation of video-language understanding in both single and multi-label settings.

We conducted a comprehensive evaluation of eight state-of-the-art VLMs under zero-shot conditions and found that current models struggle to generalize to the complexity of industrial and safety-critical tasks. Performance consistently declines on hazardous actions compared to routine activities, and models exhibit higher accuracy in multi-label settings where partial credit is possible. Moreover, models tend to perform better on object-centric, visually distinct actions than on subtle, interaction-heavy behaviors—highlighting a key shortcoming in current model reasoning capabilities. 

These findings reveal the significant gap between existing models and the demands of real-world industrial and safety applications. By offering a challenging, diverse, and safety-relevant benchmark, iSafetyBench provides a crucial testbed for driving progress in VLMs toward robust understanding in high-stakes environments.

\newpage

\section{Acknowledgement}
\label{sec:acknowledgement}
This research was supported in part by The Florida High Tech Corridor’s Matching Grant Research Program at UCF.
{
    \small
    \bibliographystyle{ieeenat_fullname}
    \bibliography{main}
}
\clearpage
\maketitlesupplementary
\renewcommand{\thesection}{\Alph{section}}
\renewcommand{\thesubsection}{\thesection.\arabic{subsection}}
\setcounter{section}{0}
\setcounter{page}{1}

In the supplementary material, we provide the detail list of actions for both the normal and hazard dataset.

\section{List of Normal Actions}

\begin{description}[leftmargin=0pt]
  \item[\textbf{Material Handling \& Movement}]\mbox{}
    \begin{enumerate}[label=\arabic*.,ref=\arabic*]
      \item pouring something
      \item holding object
      \item moving object
      \item placing object
      \item stacking objects
      \item pulling object
      \item pushing object
      \item rotating object
      \item folding fabric
      \item tying fabric
      \item stamping object
      \item loading materials
      \item unloading materials
      \item picking object
      \item rotating wood planks
      \item tearing something
      \item poking object
      \item arranging objects
      \item rolling object
      \item wrapping chain
      \item attaching label
    \end{enumerate}

  \item[\textbf{Assembly \& Production Tasks}]\mbox{}
    \begin{enumerate}[label=\arabic*.,resume]
      \item assembling electronic components
      \item screwing parts
      \item adjusting wiring
      \item inserting screws
      \item cutting with machine
      \item bending metal sheet
      \item cleaning object
      \item sealing boxes
      \item scanning label
      \item sewing
      \item connecting pipes
      \item testing product functionality
      \item heating metal
      \item opening panel cover
      \item closing panel cover
      \item digging
    \end{enumerate}

  \item[\textbf{Machinery Operation \& Maintenance}]\mbox{}
    \begin{enumerate}[label=\arabic*.,resume]
      \item operating machine
      \item pressing button
      \item adjusting fastener with wrench
      \item hammering component
      \item using jack to lift vehicle
      \item adjusting machine components
      \item detaching part from machinery
    \end{enumerate}

  \item[\textbf{Vehicle Interaction \& Site Logistics}]\mbox{}
    \begin{enumerate}[label=\arabic*.,resume]
      \item driving vehicle
      \item parking vehicle
      \item backing vehicle
      \item opening vehicle door
      \item closing vehicle door
      \item entering vehicle
      \item exiting vehicle
      \item loading to truck
      \item unloading from truck
      \item attaching chain to vehicle
      \item changing tire
      \item operating industrial vehicle
    \end{enumerate}

  \item[\textbf{Inspection \& Quality Control}]\mbox{}
    \begin{enumerate}[label=\arabic*.,resume]
      \item inspecting equipment or text
      \item looking at screen
      \item measuring with tape
      \item writing notes
      \item taking a photo
    \end{enumerate}

  \item[\textbf{Communication \& Human Interaction}]\mbox{}
    \begin{enumerate}[label=\arabic*.,resume]
      \item talking
      \item pointing at something
      \item signaling to someone
      \item posing for camera
      \item calling for help
      \item assisting injured coworker
      \item arguing
      \item arresting person
      \item knocking on window
      \item shaking hands
    \end{enumerate}

  \item[\textbf{Navigation \& Personal Movement}]\mbox{}
    \begin{enumerate}[label=\arabic*.,resume]
      \item adjusting clothing
      \item standing
      \item walking
      \item running
      \item entering doorway
      \item exiting doorway
      \item crouching
      \item bending down
      \item balancing on beam
      \item dancing
      \item showing discomfort
      \item jumping
      \item rubbing eyes
      \item rubbing head
      \item yawning
      \item checking self
      \item sitting
      \item falling down
      \item eating
      \item getting up
      \item putting on PPE
      \item taking off PPE
      \item crawling
    \end{enumerate}

  \item[\textbf{Surveillance \& Observation}]\mbox{}
    \begin{enumerate}[label=\arabic*.,resume]
      \item monitoring onboard camera
      \item appearing on security feed
      \item searching area with flashlight
      \item watching process
    \end{enumerate}
\end{description}

\section{List of Dangerous/Hazard Actions}
\begin{description}[leftmargin=0pt]
  \item[\textbf{Machinery \& Equipment Operation Errors}]\mbox{}
    \begin{enumerate}[label=\arabic*.,ref=\arabic*]
      \item operating forklift
      \item forklift blade detaching
      \item operating heavy equipment dangerously
      \item operating hydraulic press
      \item operating rolling machine
      \item rotating machine lever
      \item adjusting machine while running
      \item robotic arm misoperation
      \item walking on moving conveyor
      \item misusing lift platform
      \item machine part flying off
      \item improper use of flamethrower
    \end{enumerate}

  \item[\textbf{Entanglement \& Compression Hazards}]\mbox{}
    \begin{enumerate}[label=\arabic*.,resume]
      \item shirt caught in machine
      \item hair caught in appliance
      \item foot stuck in conveyor
      \item body pulled into machine
      \item trapped between closing machine sides
      \item crushed under overturned vehicle
    \end{enumerate}

  \item[\textbf{Structural Failures, Falling Objects \& Collapses}]\mbox{}
    \begin{enumerate}[label=\arabic*.,resume]
      \item structural collapse
      \item falling load
      \item heavy object slipping
      \item break under load
      \item warehouse shelves toppling
      \item crane imbalance with suspended load
      \item glass shattering on impact
      \item unloading truck load fall
    \end{enumerate}

  \item[\textbf{Fire, Explosion \& Thermal Hazards}]\mbox{}
    \begin{enumerate}[label=\arabic*.,resume]
      \item fire incident
      \item pressurized vessel explosion
      \item machine explosion
      \item gas burst
    \end{enumerate}

  \item[\textbf{Manual Handling \& Lifting Incidents}]\mbox{}
    \begin{enumerate}[label=\arabic*.,resume]
      \item lifting heavy load
      \item pushing heavy load
      \item pulling heavy load
      \item carrying heavy load
      \item carrying object and slipping
    \end{enumerate}

  \item[\textbf{Vehicle \& Traffic-Related Incidents}]\mbox{}
    \begin{enumerate}[label=\arabic*.,resume]
      \item driving car
      \item vehicle crash into building or stationary object
      \item vehicle losing control
      \item collision between vehicles
      \item driver thrown during crash
      \item car hood malfunction or improper interaction
      \item person dragged by vehicle
      \item vehicle falling off edge
    \end{enumerate}

  \item[\textbf{Rescue, Situational Awareness \& Human Error}]\mbox{}
    \begin{enumerate}[label=\arabic*.,resume]
      \item lifting object off person
      \item extinguishing fire
      \item rescue effort
      \item helping person down from height
      \item searching debris for victims
      \item signalling others about hazard
      \item filming incident
      \item watching incident passively
      \item escaping from danger
    \end{enumerate}

  \item[\textbf{Environmental \& Workplace Hazards}]\mbox{}
    \begin{enumerate}[label=\arabic*.,resume]
      \item equipment emitting sparks
      \item overstacked shelves
      \item cluttered workspace
      \item flooded floor
      \item platform failure
    \end{enumerate}

  \item[\textbf{Slips, Trips, Falls \& Non-Machine Events}]\mbox{}
    \begin{enumerate}[label=\arabic*.,resume]
      \item person falling down
      \item tree falling nearby
      \item hanging from something after slip
    \end{enumerate}

  \item[\textbf{Security Incidents \& Intentional Misconduct}]\mbox{}
    \begin{enumerate}[label=\arabic*.,resume]
      \item moving in a suspicious manner
      \item physical altercation or fighting
      \item vandalism or intentional property damage
      \item intention of theft
      \item police arrest
      \item police search
      \item firearm discharge
    \end{enumerate}
\end{description}

\end{document}